\RequirePackage{fix-cm}
\documentclass[smallextended]{svjour3}       
\smartqed  

\usepackage{graphicx}
\usepackage{mathptmx}
\usepackage{algorithm}
\usepackage{booktabs} 
\usepackage{algorithmic}
\usepackage{graphicx}
\usepackage{multirow}
\usepackage{amsmath}
\usepackage{xcolor}
\usepackage{color}
\begin{document}

\title{A multilayer backpropagation saliency detection algorithm and its applications
}


\author{Chunbiao Zhu and Ge Li
}

\institute{ Chunbiao Zhu \at
              \email{zhuchunbiao@pku.edu.cn}            
           \and
           Ge Li \at
               \email{geli@ece.pku.edu.cn}
           \and
              Prof.Li is the corresponding author. The authors are with the School of Electronic and Computer Engineering, Peking University Shenzhen Graduate School, Shenzhen 518055, China\\
              \\
              This work was supported by the grant of National Natural Science Foundation of China (No.U1611461), the grant of Science and Technology Planning Project of Guangdong Province, China (No.2014B090910001) and the grant of Shenzhen Peacock Plan (No.20130408-183003656).
}

\date{Received: date / Accepted: date}

\maketitle

\begin{abstract}
Saliency detection is an active topic in the multimedia field.
 Most previous works on saliency detection focus on 2D images. However, these methods are not robust against complex scenes which contain multiple objects or complex backgrounds.
 Recently, depth information supplies a powerful cue for saliency detection.
 In this paper, we propose a multilayer backpropagation saliency detection algorithm based on depth mining by which we exploit depth cue from three different layers of images.
 The proposed algorithm shows a good performance and maintains the robustness in complex situations. Experiments’ results show that the proposed framework is superior to other existing saliency approaches.
 Besides, we give two innovative applications by this algorithm, such as scene reconstruction from multiple images and small target object detection in video.
\keywords{Saliency Detection \and Multilayer \and Backpropagation \and Depth Mining \and Image Montage \and Small Target Detection}

\end{abstract}

\section{Introduction}
\label{intro}
Salient object detection is a process of getting the visual attention region precisely.
 The attention is the behavioral and cognitive process of selectively concentrating on one aspect of the environment while ignoring other things.

Early work on computing saliency aims to locate the visual attention region. Recently the field has been extended to locate and refine the salient regions and objects.
 Served as a foundation of various multimedia applications, salient object detection has been widely used in content-aware editing~\cite{Chang2011Content}, image retrieval~\cite{Cheng2014SalientShape}, object recognition~\cite{Alexe2012Measuring}, object segmentation~\cite{Girshick2013Rich}, compression~\cite{Itti2004Automatic}, image retargeting~\cite{Sun2011Scale}, etc.

Generally speaking, saliency detection frameworks mainly use top-down or bottom-up approaches.
 Top-down approaches are task-driven and need supervised learning.
 While bottom-up approaches usually use low-level cues, such as color features, distance features, depth features and heuristic saliency features.

 The most used features are heuristic saliency features and discriminative saliency features. Various measures based on heuristic saliency features have been proposed, including pixel-based or patch-based contrast~\cite{Ma2003Contrast,Liu2006Region,Achanta2008Salient,Valenti2009Image}, region-based contrast~\cite{Cheng2011Global,Jiang2011Automatic,Krahenbuhl2012Saliency,Li2013Contextual,Jiang2013Salient,Ran2013What,Shi2013PISA,Li2013Estimating}, pseudo-background~\cite{Wei2012Geodesic,Jiang2013Saliency,Yang2013Saliency,Li2013Saliency,Liu2014Adaptive,Zhu2014Saliency}, and similar images~\cite{Marchesotti2009A,Siva2013Looking}. Some measures proposed use discriminative saliency features, such as multi-scale contrast~\cite{Liu2011Learning}, center-surround contrast~\cite{Jiang2013Salient}, and color spatial compactness~\cite{Mehrani2010Saliency}. And other measures use image over-segmentation~\cite{Cheng2014Efficient}, outlier~\cite{Wu2012A,Xie2013Bayesian,Lu2011Salient,Chang2011Fusing}， wavelet features~\cite{Imamoglu2013A,You2010A}, which can provide multi-scale spatial and frequency analysis at the same time for saliency detection, and other features representation~\cite{Sun2017An,Zhao2017Continuous}.

Although these methods can make full use of 1-2 RGB-based features, they are not robust to specific situations and lead to the inaccuracy of results on challenging salient object detection datasets.

Recently, advances in 3D data acquisition techniques have motivated the adoption of
structural features, improving the discrimination between different objects with the similar appearance.
some algorithms~\cite{Zhu2017Salient,Peng2014RGBD,Cheng2014Depth,Geng2012Leveraging,zhu2017innovative,zhu2017three} adopt depth cue to deal with the challenging scenarios.
In~\cite{Zhu2017Salient}, Zhu et al. propose a framework based on cognitive neuroscience, and use depth cue to represent the depth of real field. In~\cite{Cheng2014Depth}, Cheng et al. compute salient stimuli in both color and depth spaces. In~\cite{Peng2014RGBD}, Peng et al. provide a simple fusion framework that combines existing RGB-based saliency with new depth-based saliency. In~\cite{Geng2012Leveraging}, Geng et al. define saliency using depth cue computed from stereo images. Their results show that stereo saliency is a useful consideration compared to previous visual saliency analysis. All of them demonstrate the effectivity of depth cue in the improvement of salient object detection.

However, depth cue cannot make the saliency results robust when a salient object has
low depth contrast compared to the background. So, how to integrate depth cue with RGB information is the problem that needs to be solved.

In this paper, we propose a multilayer backpropagation saliency detection algorithm based on depth mining~\cite{zhu2017multilayer} to overcome the aforementioned problem.
First, we obtain the center-bias saliency map and depth map in the preprocessing stage.
Then, we dispose of the input images in three different layers, respectively. In the first layer, we use original depth cue and other cues to calculate preliminary saliency value.
In the second layer, we apply processed depth cue and other cues to compute intermediate saliency value.
In the third layer, we employ reprocessed depth cue and other cues to get final saliency value. The framework of the proposed algorithm is illustrated in Fig. 1.
The experiments show that the proposed algorithm is both effective and robustness in saliency detection.
Besides, we give two innovative applications to demonstrate the potential of saliency detection for broad applications in computer vision and computer graphics.

In summary, the main contributions of our work include:

1. We propose a multilayer backpropagation saliency detection algorithm based on depth mining, which has a good performance in saliency detection.

2. We use the proposed algorithm to the application of image montage, and the constructed image montages’ exquisiteness illustrates that our method is outperformance.

3. We employ a novel approach to the small target detection application, the detection results are superb.

 In addition, although saliency detection has recently attracted much attention, its practical usage for real vision tasks has yet to be justified. Our method validates the usefulness of saliency detection by implementing two applications.

The rest of the paper is organized as follows: we introduce the multilayer backpropagation saliency detection algorithm in Sect. 2.
In Sect. 3, we show the experimental results of the proposed algorithm.
In Sect. 4, we give an application of image montage to reconstruct scenes from multiple images.
We further show a novel approach to the application of small target detection in Sect. 5. At last, we
conclude the paper in Sect. 6.
\begin{figure}[!htb]
\label{fig:1}
\begin{center}
\includegraphics[width=4.7in,height =1.62in]{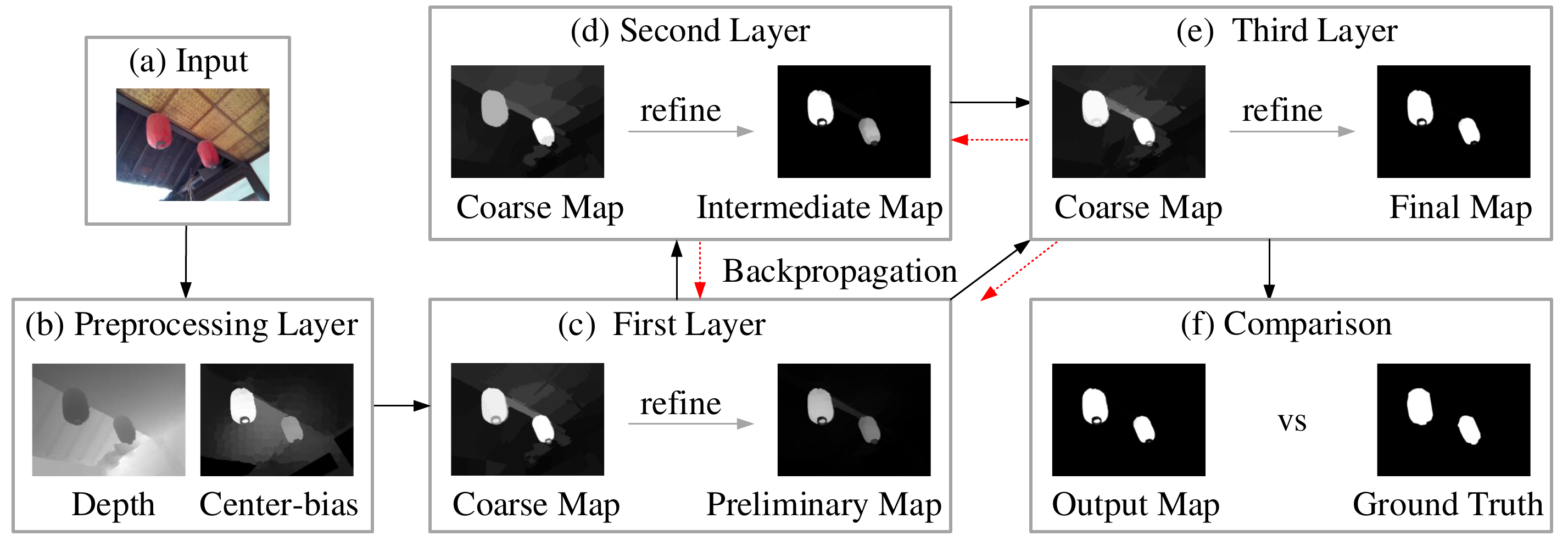}
\end{center}
\caption{The framework of the proposed algorithm.}
\label{fig:short}
\end{figure}
\section{Proposed Algorithm}
\label{sec:2}
As shown in Fig. 1, the framework of the proposed algorithm contains four layers, including the preprocessing layer, the first layer, the second layer and the third layer.
\subsection{The Preprocessing Layer}
\begin{figure}[!htb]
\label{fig:2}
\begin{center}
\includegraphics[width=4.5in,height =1.3in]{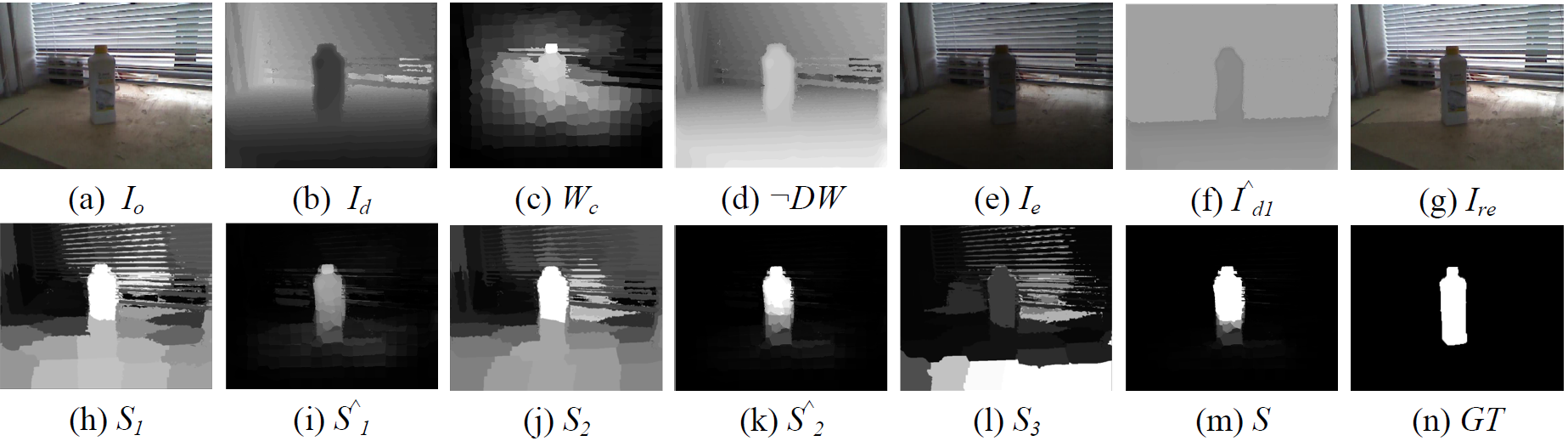}
\end{center}
\caption{The visual process of the proposed algorithm.}
\label{fig:short}
\end{figure}
\label{sec:2.1}
In the preprocessing layer, we imitate the human eyes to obtain center-bias saliency map and depth map.
\paragraph{Center-bias Saliency Map.} Inspired by cognitive neuroscience,
human eyes use central fovea to locate objects and make them
clearly visible. Therefore, most of the images taken by cameras
always locate salient object around the center. Aiming to get
center-bias saliency map, we use BSCA algorithm~\cite{Qin2015Saliency}. It
constructs global color distinction and spatial distance matrix
based on clustered boundary seeds and integrates them into a
background-based map. Thus it can improve the center-bias,
erasing the image edge effect. As shown in the preprocessing
stage of Fig. 2 (c), the center-bias saliency map can remove the
surroundings of the image and reserve most of the salient regions.
We denote this center-bias saliency map as $W_{c}$.
\paragraph{Depth Map.} Similarly, biology prompting shows that people perceive the distance and the depth of the object mainly relies on two eyes, which provide the clues, and we call it binocular parallax.
Therefore, the depth cue can imitate the depth of real field. The depth map used in the experimental datasets is taken by Kinect device. And we denote the depth map as $I_{d}$.
\subsection{The First Layer}
\label{sec:2.2}
In the first layer, we extract color and depth features from the original image $I_{o}$ and the depth map $I_{d}$, respectively.

First, the image $I_{o}$ is segmented into $K$ regions based on color via the $K$-means algorithm. Define:
\begin{equation}
  S_{c}(r_{k}) =\sum_{i=1,i\neq k}^K P_i W_d(r_k) D_c(r_k,r_i),
\end{equation}
where $S_c(r_k )$ is the color saliency of region $k$, $k\in[1,K]$, $r_k$ and $r_i$ represent regions $k$ and $i$ respectively, $D_c (r_k,r_i )$ is the Euclidean distance between region $k$ and region $i$ in L*a*b color space, $P_i$ represents the area ratio of region $r_i$ compared with the whole image, $W_d (r_k )$ is the spatial weighted term of the region $k$, set as:
\begin{equation}
  W_d (r_k )=e^{\frac{-D_o (r_k,r_i)}{\sigma^2}},
\end{equation}
where $D_o (r_k,r_i)$ is the Euclidean distance between the centers of region $k$ and $i$, $\sigma$ is the parameter controlling the strength of $W_d (r_k)$.

Similar to color saliency, we define:
\begin{equation}
 S_d (r_k )=\sum_{i=1,i\neq k}^K P_i W_d (r_k ) D_d (r_k,r_i ),
\end{equation}
where $S_d (r_k)$ is the depth saliency of $I_d$, $D_d (r_k,r_i)$ is the Euclidean distance between region $k$ and region $i$ in depth space.

In most cases, a salient object always locate at the centre of an image or close to a camera. Therefore, we assign the weights to both centre-bias and depth for both color and depth images. The weight of the region k is set as:
\begin{equation}
W_{cd} (r_k)=\frac{G(‖P_k-P_o ‖)}{N_k}DW(d_k ),
\end{equation}
where $G(\cdot)$ represents the Gaussian normalization, $\|\cdot\|$ is Euclidean distance, $P_k$ is the position of the region $k$, $P_o$  is the center position of this map, $N_k$ is the number of pixels in region $k$, $DW(d_k)$ is the depth weight, which is set as:
\begin{equation}
DW(d_k )=(max\{d\}-d_k )^\mu ,
\end{equation}
where $max\{d\}$ represents the maximum depth of the image, and $d_k$ is the depth value of region $k$, $μ$ is a fixed value for a depth map, set as:
\begin{equation}
\mu=\frac{1}{max\{d\}-min\{d\}},
\end{equation}
where $min\{d\}$ represents the minimum depth of the image.

Second, the preliminary saliency value of the region $k$ is calculated as:
\begin{equation}
S_1 (r_k )=G(S_c (r_k )\times W_{cd} (r_k )+S_d (r_k )\times W_{cd} (r_k )),
\end{equation}

Third, to get a refinement results, we refine the preliminary saliency map with the help of the center-bias and depth maps.
The final preliminary saliency value is calculated as following:
\begin{equation}
\hat{S_1}(r_k)=S_1 (r_k)\times \neg DW(d_k )\times W_{c}.
\end{equation}
where $\neg$ is the negation operation which can enhance the saliency degree of front regions as shown in Fig. 2(d), because the foreground object has low depth value in depth map while the background object has high depth value.
$W_{c}$ is the center-bias saliency value calculated in the preprocessing layer, which can improve the center-bias, erasing the image edge effect as shown in Fig. 2(c).

\begin{algorithm}[!htb]
\label{alg:1}
\caption{ Procedure for the first layer}
\hspace*{0.02in} {\bf Input:}
original maps $I_o$, center-bias map$W_c$,depth maps $I_d$;\\
\hspace*{0.02in} {\bf Output:}
the final preliminary saliency values $\hat{S_1}(r_k)$;
\begin{algorithmic}[1]
\STATE {\bf for} each region $k=1,K$ {\bf do:}
\STATE compute color saliency values $S_c (r_k )$ and depth saliency values $S_d (r_k )$;
\STATE calculate the center-bias and depth weights $W_{cd} (r_k )$;
\STATE get the preliminary saliency values $S_1 (r_k )$;
\STATE calculate the final preliminary saliency values $\hat{S_1}(r_k)$;
\STATE {\bf end for}
\RETURN the final preliminary saliency values $\hat{S_1}(r_k)$.
\end{algorithmic}
\end{algorithm}
\subsection{The Second Layer}
\label{sec:2.3}
In the second layer, first, we set:
\begin{equation}
I_e\{R\mid G\mid B\} =I_o\{R\mid G\mid B\} \times I_d,
\end{equation}
where $I_e$ represents the extended map. $\{R\mid G\mid B\}$ represents processing of three RGB channels, respectively.

The extended map is displayed in Fig. 2(e), from which the salient objects’ edges are prominent.

Second, we use extended map $I_e$ to replace $I_o$. Then, we calculate intermediate saliency value via the first layer’ via Eq. 7. We get:
\begin{equation}
S_2 (r_k )=G(S_c (r_k )\times W_{cd} (r_k )+S_d (r_k )\times W_{cd} (r_k )),
\end{equation}
where $S_2 (r_k )$ is the intermediate saliency value.

Third, to refine intermediate saliency value, we apply the backpropagation to enhance the intermediate saliency value by mixing the result of the first layer.
And we define our final intermediate saliency value as:
\begin{equation}
\hat{S_2}(r_k)=\hat{S_1}^{2}(r_k)+\hat{S_1}(r_k)\times (1-e^{-S_2^{2}(r_k )\times \neg DW(d_k )}).
\end{equation}

\begin{algorithm}[!htb]
\label{alg:2}
\caption{ Procedure for the second layer}
\hspace*{0.02in} {\bf Input:}
extended map $I_e$, depth maps $I_d$;\\
\hspace*{0.02in} {\bf Output:}
the final intermediate saliency values $\hat{S_2}(r_k)$;
\begin{algorithmic}[1]
\STATE {\bf for} each region $k=1,K$ {\bf do:}
\STATE compute color saliency values $S_c (r_k )$ and depth saliency values $S_d (r_k )$;
\STATE calculate the center-bias and depth weights $W_{cd} (r_k )$;
\STATE get the intermediate saliency value $S_1 (r_k )$;
\STATE calculate the final intermediate saliency values $\hat{S_2}(r_k)$;
\STATE {\bf end for}
\RETURN the final intermediate saliency values $\hat{S_2}(r_k)$.
\end{algorithmic}
\end{algorithm}
\subsection{The Third Layer}
\label{sec:2.4}
In the third layer, first, we reprocess the depth cue by filtering the depth map via the following formula:
\begin{equation}
I_{d1}=
 \begin{cases}
I_{d1}, d\leq \beta \times max\{d\} \\
0     , d >\beta \times max\{d\}
  \end{cases}
,
\end{equation}
where $I_{d1}$ represents the filtered depth map. $\beta$ is the parameter which controls the length of  $I_{d1}$.
In general, salient objects always have the small depth value compared to background, thus, by Eq. 12, we can filter out the background noises.

Second, we polarize the filtered depth map shown in Fig. 2(g) via the following formula:
\begin{equation}
\hat{I_{d1}}=1-e^{-(1-I_{d1})},
\end{equation}

Third, we extend the filtered depth map to the color images via the Eq. 9. We denote the reprocessed depth map as $I_{re}$

We use filtered depth map $I_{re}$ to replace $I_o$. Then, by Eq. 7, we get the third layer saliency value denoted as:
\begin{equation}
S_3 (r_k )=G(S_c (r_k )\times W_{cd} (r_k )+S_d (r_k )\times W_{cd} (r_k )),
\end{equation}

Fourth, to refine $S_3 (r_k )$, we apply the backpropagation of $\hat{S_1}(r_k)$ and $\hat{S_2}(r_k)$  as following:
\begin{equation}
{S}(r_k)=\hat{S_2}(r_k)\times (\hat{S_1}(r_k)+S_3 (r_k ))\times (S_3 (r_k )+1-e^{-\hat{S_1}(r_k )\times {S_3}^{2}(r_k) }).
\end{equation}

From the Fig. 2, we can see the visual results of the proposed algorithm. The main steps of the proposed salient object detection algorithm are summarized in Algorithm 1.
\begin{algorithm}[!htb]
\label{alg:3}
\caption{ Procedure for the third layer}
\hspace*{0.02in} {\bf Input:}
extended map $I_{re}$, depth maps $I_d$;\\
\hspace*{0.02in} {\bf Output:}
the final saliency values $S(r_k)$;
\begin{algorithmic}[1]
\STATE {\bf for} each region $k=1,K$ {\bf do:}
\STATE compute color saliency values $S_c (r_k )$ and depth saliency values $S_d (r_k )$;
\STATE calculate the center-bias and depth weights $W_{cd} (r_k )$;
\STATE get the intermediate saliency value $S_1 (r_k )$;
\STATE calculate the final saliency values $S(r_k)$;
\STATE {\bf end for}
\RETURN the final saliency values $S(r_k)$.
\end{algorithmic}
\end{algorithm}
%
\section{Experiments}
\label{sec:3}
\subsection{Datasets and Evaluation Indicators}
\label{sec:3.1}
\paragraph{Datasets.} We evaluate the performance of the proposed saliency detection algorithm on two RGBD standard datasets: RGBD1*~\cite{Cheng2014Depth} and RGBD2*~\cite{Peng2014RGBD}. RGBD1* has 135 indoor images taken by Kinect with the resolution of $640\times480$. This dataset has complex backgrounds and irregular shapes of salient objects. RGBD2* contains 1000 images with both indoor and outdoor images.
\paragraph{Evaluation indicators.} Experimental evaluations are based on standard measurements
including precision-recall curve, ROC curve, MAE (Mean Absolute Error), F-measure, Max-P(the maximum value of precision), Min-P(the minimum value of precision), Max-R(the maximum value of recall), Min-R(the minimum value of recall). Among them, the MAE is formulated as:
\begin{equation}
MAE=\frac{\sum_{i=1}^N \|GT_i-S_i\|}{N}.
\end{equation}
where $N$ is the number of the testing images, $GT_i$ is the area of the ground truth of image $i$, $S_i$ is the area of detection result of image $i$.

And the F-measure is formulated as:
\begin{equation}
F-measure=\frac{2\times Precision\times Recall}{Precision+Recall}.
\end{equation}
\subsection{Ablation Study}
\label{sec:3.2}
We first validate the effectiveness of each step in our method: the first step results, the second step results and the third step results. Table. 1 shows the validation results on two datasets. We can clear see the accumulated processing gains after each step, and the final saliency results shows a good performance. After all, it proves that each steps in our algorithm is effective for generating the final saliency maps.
\begin{table}[!hbp]
\caption{The validation results of each step in the proposed algorithm.}
\label{tab:1}
\begin{tabular}{|p{2.5in}|p{0.5in}|p{0.5in}|p{0.5in}|}
\hline
Each Steps of the Proposed Algorithm & $\hat{S_1}(r_k)$ & $\hat{S_2}(r_k)$ & $S(r_k)$ \\
\hline
MAE Values on RGBD1* Dataset &0.1065 &0.0880&0.0781\\
\hline
MAE Values on RGBD2* Dataset &0.1043 &0.0900&0.0852\\
\hline
F-measure Values on RGBD1* Dataset &0.5357 &0.6881&0.7230\\
\hline
F-measure Values on RGBD2* Dataset &0.5452 &0.7025&0.7190\\
\hline
\end{tabular}
\end{table}
\begin{figure}[!htb]
\label{fig:3}
\begin{center}
\includegraphics[width=4.7 in,height =1.12 in]{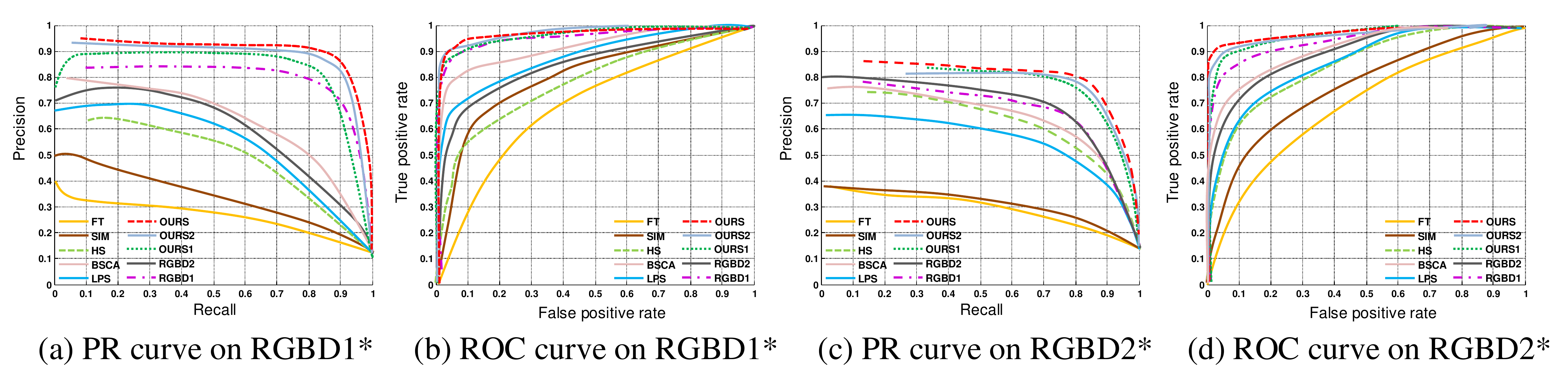}
\end{center}
\caption{ (a): PR curve of different methods on RGBD1* dataset. (b): ROC curve of different methods on RGBD1* dataset.(c): PR curve of different methods on RGBD1* dataset. (d): ROC curve of different methods on RGBD2* dataset.  }
\label{fig:short}
\end{figure}
\begin{table}[!htb]
\label{tab:2}
\caption{The evaluation results on RGBD1* dataset, the best results are shown in boldface.}
\begin{tabular}{|p{0.5in}|p{0.5in}|p{0.5in}|p{0.5in}|p{0.5in}|p{0.5in}|p{0.5in}|}
\hline
Methods & MAE & F-measure & Max-P & Min-P & Max-R & Min-R \\
\hline
FT &0.2049 & 0.2804 & 0.3875 & \textbf{0.1262} & \textbf{1} & 0\\
\hline
SIM &0.3740 & 0.3345 & 0.5076 & \textbf{0.1262} & \textbf{1} & 0\\
\hline
HS &0.1849 & 0.5361 & 0.6581 & \textbf{0.1262} & \textbf{1} & 0.1187\\
\hline
BSCA &0.1851 & 0.5826 & 0.7977 & \textbf{0.1262} & \textbf{1} & 0.0306\\
\hline
LPS &0.1406 & 0.5452 & 0.6951 & \textbf{0.1262} & \textbf{1} & 0.0026\\
\hline
RGBD1 &0.3079 & 0.5410 & 0.8561 & \textbf{0.1262} & \textbf{1} & 0.1731\\
\hline
RGBD2 &0.1165 & 0.4912 & 0.7699 & \textbf{0.1262} & \textbf{1} & 0.0049\\
\hline
OURS1 & 0.1065 & 0.5357 & 0.9000 & 0.0074 & \textbf{1} & 0\\
\hline
OURS2 & 0.0880 & 0.6881 & \textbf{0.9249} & \textbf{0.1262} & \textbf{1} & 0.1118\\
\hline
OURS & \textbf{0.0781} & \textbf{0.7230} & 0.9181 & \textbf{0.1262} & \textbf{1} & \textbf{0.2669}\\
\hline
\end{tabular}
\end{table}
\begin{table}[!htb]
\label{tab:3}
\caption{The evaluation results on RGBD2* dataset, the best results are shown in boldface.}
\begin{tabular}{|p{0.5in}|p{0.5in}|p{0.5in}|p{0.5in}|p{0.5in}|p{0.5in}|p{0.5in}|}
\hline
Methods & MAE & F-measure & Max-P & Min-P & Max-R & Min-R \\
\hline
FT &0.2168 & 0.3270 & 0.3894 & \textbf{0.1291} & \textbf{1} & 0\\
\hline
SIM &0.3957 & 0.2927 & 0.3847 & \textbf{0.1291} & \textbf{1} & 0\\
\hline
HS &0.1909 & 0.6003 & 0.7503 & \textbf{0.1291} & \textbf{1} & 0.1859\\
\hline
BSCA &0.1754 & 0.5925 & 0.7616 & \textbf{0.1291} & \textbf{1} & 0.0525\\
\hline
LPS &0.1252 & 0.5890 & 0.6831 & \textbf{0.1291} & \textbf{1} & 0.0166\\
\hline
RGBD1 &0.3207 & 0.4843 & 0.7771 & \textbf{0.1291} & \textbf{1} & 0.2228\\
\hline
RGBD2 &0.1087 & 0.5957 & 0.8148 & \textbf{0.1291} & \textbf{1} & 0.0070\\
\hline
OURS1 & 0.1043 & 0.5452 & 0.8029 & 0.0150 & \textbf{1} & 0\\
\hline
OURS2 & 0.0900 & 0.7025 & \textbf{0.8477} & \textbf{0.1291} & \textbf{1} & 0.2445\\
\hline
OURS & \textbf{0.0852} & \textbf{0.7190} & 0.8347 & \textbf{0.1291} & \textbf{1} & \textbf{0.4071}\\
\hline
\end{tabular}
\end{table}
\begin{figure}
\label{fig:4}
\begin{center}
\includegraphics[width=4.882 in,height = 6.37 in]{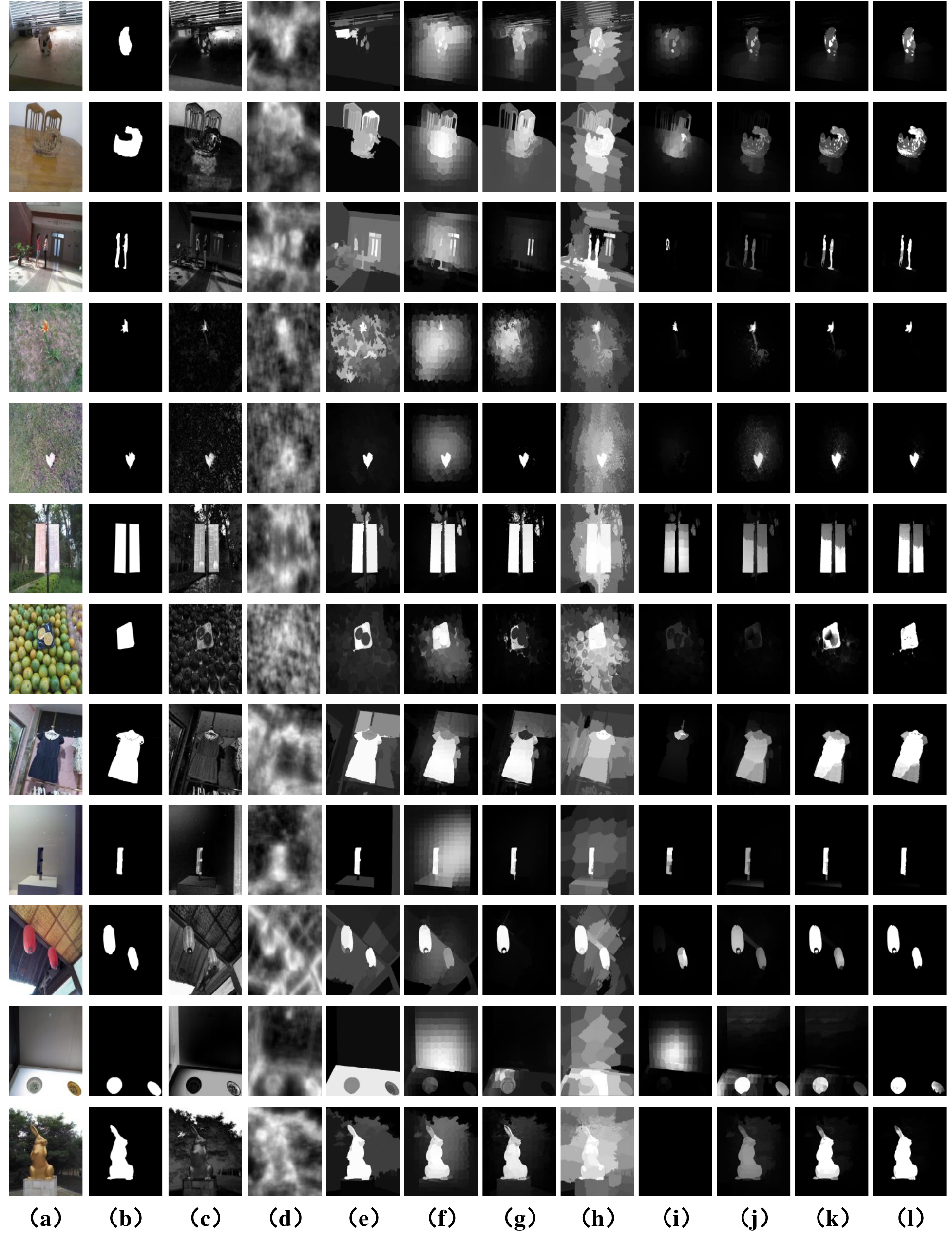}
\end{center}
\caption{Visual comparison of saliency maps on two datasets. (a) - (l) represent: input images, ground truth, FT, SIM, HS, BSCA, LPS, RGBD1, RGBD2, OURS1, OURS2 and OURS, respectively.}
\label{fig:short}
\end{figure}
\subsection{Comparison}
To further illustrate the effectiveness of our algorithm, we compare our proposed methods with FT~\cite{Achanta2009Frequency}, SIM~\cite{Murray2011Saliency}, HS~\cite{Shi2016Hierarchical}, BSCA~\cite{Qin2015Saliency}, LPS~\cite{Li2015Inner}, RGBD1~\cite{Cheng2014Depth}, RGBD2~\cite{Peng2014RGBD}. We use the codes provided by the author to reproduce their experiments. For all the compared methods, we use the default settings suggested by the authors. Besides, to show that the contribution of depth mining using multi-layers, we add the intermediate results (OURS1 and OURS2) for the first layer and second layer in the experiment comparisons. And for the Eq. 2 and Eq. 12, we take $\sigma^2 = 0.4$ and $\beta = 0.3$, respectively, which has the best contribution to the results.

The precision and recall evaluation results and ROC evaluation results are shown in Fig. 3.
From the precision-recall curves and ROC curves, we can see that our multi-layers’ saliency detection results can achieve better results on both RGBD1* and RGBD2* datasets.

Other evaluation results on both RGBD1* and RGBD2* datasets are shown in Table 2 and Table 3, respectively. The best results are shown in boldface. Compared with the MAE values, it can be observed that our saliency detection method is superior and can obtain more precise salient regions than that of other approaches. Besides, the proposed algorithm is the most robust.

The visual comparisons are given in Fig. 4, which clearly demonstrate the advantages of our method. We can see that our method can detect both single salient object and multiple salient objects more precisely. Besides, by intermediate results, it shows that by exploiting depth cue information of more layers, our proposed method can get more accurate and robust performance. In contrast, the compared methods may fail in some situations.
\section{Image Montage Application}
\label{sec:4}
In this section, we use the proposed algorithm to an innovative application of image montage.
Our image montage application is divided into six stages, including saliency detection, object segmentation, color changing, object resizing, object removal and scene reconstruction.The performance of most stages is highly dependent on salient.
\subsection{Salient Object Detection}
\label{sec:4.1}
To get the image montage, first, we gather some objects that we are interested in.
Therefore, we use the proposed algorithm to obtain those objects. Since the proposed algorithm has more detection precision regions,
the following stages will reduce errors and can achieve better visual effect of image montage. We use the proposed algorithm to get the object saliency maps shown in Fig. 5(b).
\begin{figure}
\label{fig:5}
\begin{center}
\includegraphics[width=4.6 in,height =7.912 in ]{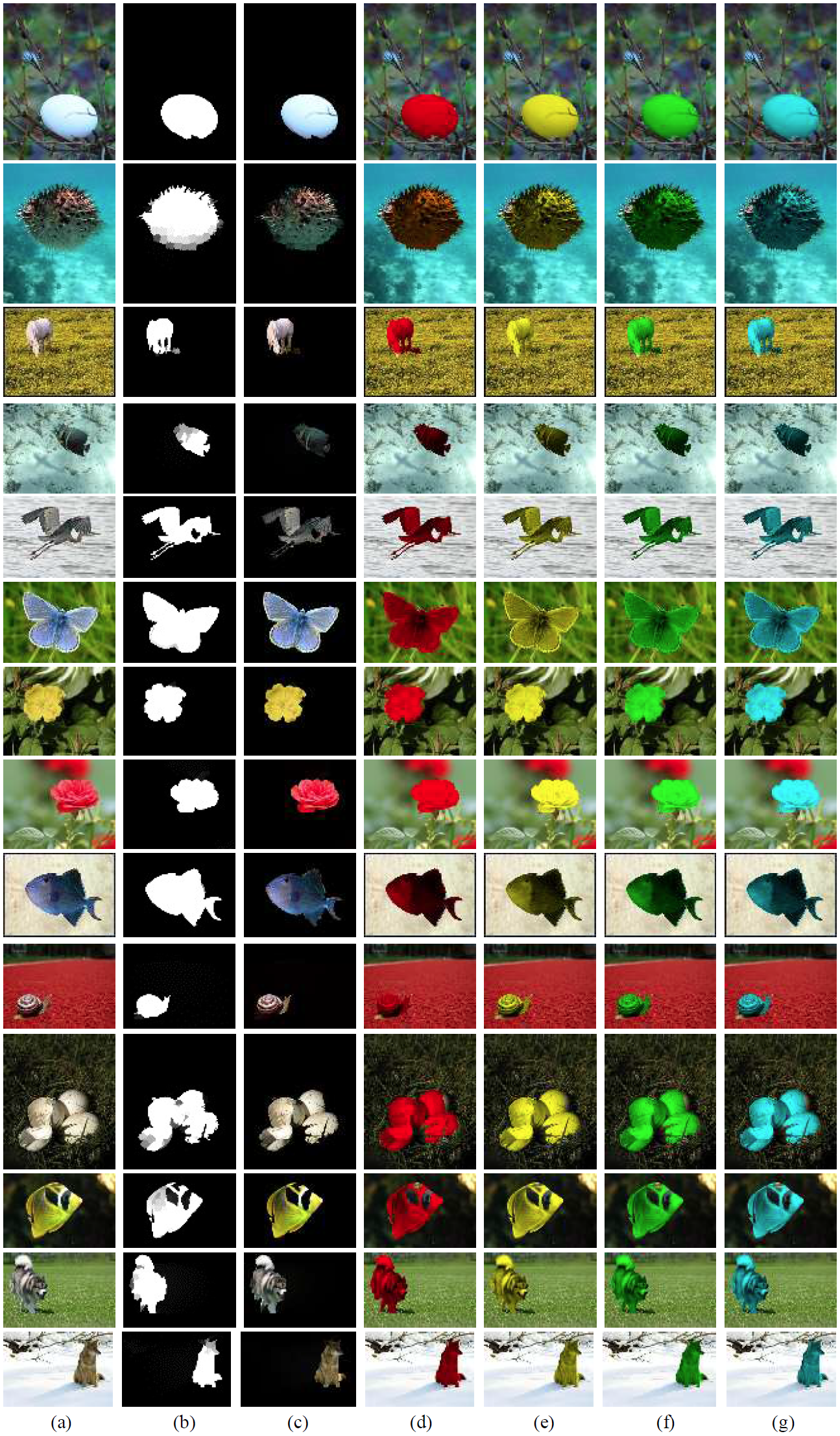}
\end{center}
\caption{Image materials. (a): original maps (b): saliency maps (c): segmentation maps (d)-(g): color changing maps.}
\label{fig:short}
\end{figure}
\subsection{Object Segmentation}
\label{sec:4.2}
After getting the object we are interested in, then we segment them from the original image scene.
In this stage, we use our salient object results and the original images’ RGB channel to recover the salient object maps into color maps. Shown as:
\begin{equation}
Seg=I_o\{R\mid G\mid B\} \times {S}.
\end{equation}
where $Seg$ represents the segmentation map. $\{R\mid G\mid B\}$ represents the processing of three RGB channels, respectively. $S$ is the saliency values calculated by the proposed algorithm.

All the results are shown in Fig. 5(c). And we can see that the more accurate salient maps we use, the more accurate segmentation results we will have.
\subsection{Color Changing}
\label{sec:4.3}
After getting the objects we are interested in, we also want to change the salient objects’ color.
Thus we use our saliency maps as the sample maps and use original maps’ RGB values to change the RGB values.
The results are shown in Fig. 5(d)-(f).
\subsection{Object Resizing}
\label{sec:4.4}
There are some situations that the salient objects are too large to fill new pictures.
Therefore, we use bilinear interpolation to resize the salient objects and the background scenes.
\begin{figure}
\label{fig:6}
\begin{center}
\includegraphics[width=4.7 in,height =2.1267 in ]{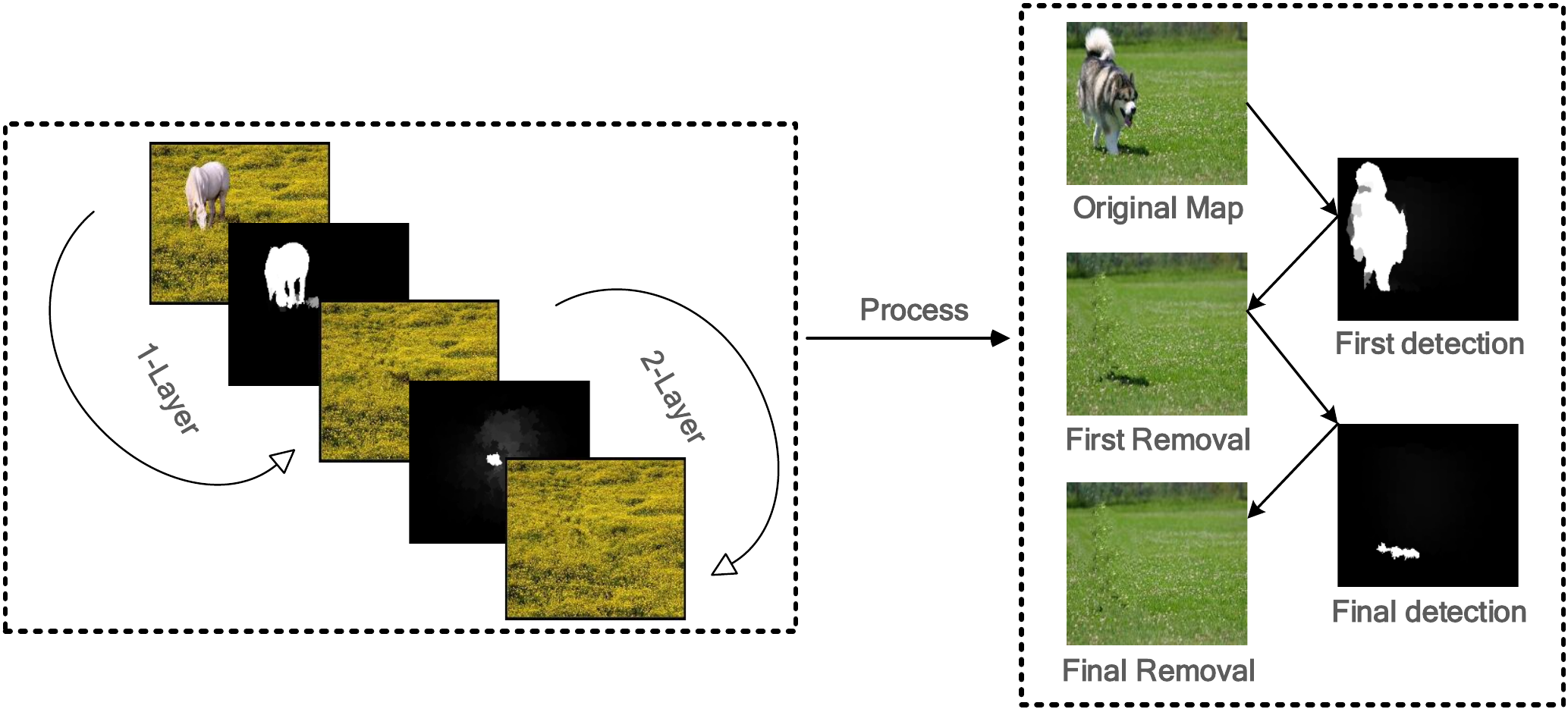}
\end{center}
\caption{Two-Layer Removal Framework.}
\label{fig:short}
\end{figure}
\begin{figure}
\label{fig:7}
\begin{center}
\includegraphics[width=4.6 in,height =1.73 in ]{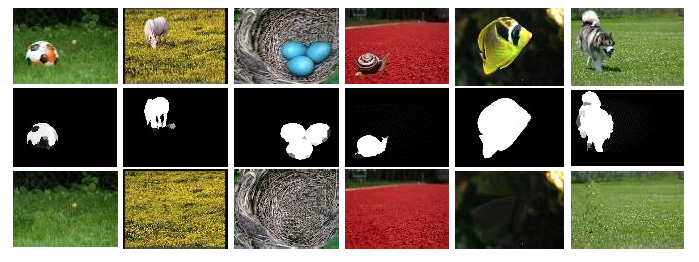}
\end{center}
\caption{Removal results. From the first line to the third line are original maps, saliency maps and removal maps, respectively.}
\label{fig:short}
\end{figure}
\subsection{Object Removal}
\label{sec:4.5}
In this stage, we want to get images’ background and remove the object that we want to change.
Inspired by the Criminisi algorithm~\cite{Criminisi2003Object}, this algorithm divided three stages.
First, the user selects the target region and then compute the patch priorities to propagate texture and structure information.
At last, updating confidence values to get the final results.
\begin{figure}
\label{fig:8}
\begin{center}
\includegraphics[width=4.7 in,height =2.956 in ]{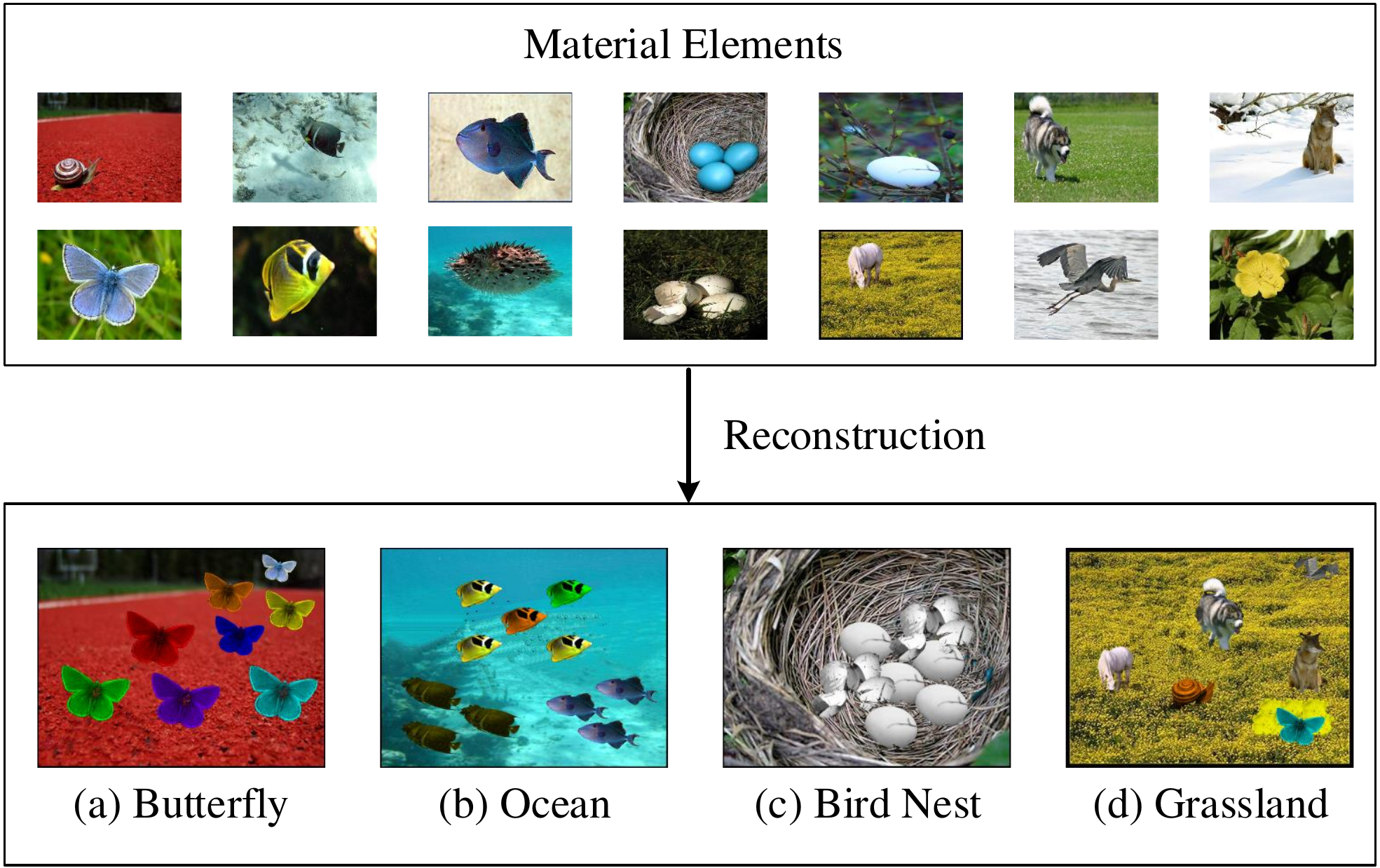}
\end{center}
\caption{Four scenarios are reconstructed by the image montage application.}
\label{fig:short}
\end{figure}
For this algorithm, we simplify the first stage which replacing the human marked regions with our salient object detection maps.
And for some objects have the shadow residues, we propose a two-layer removal algorithm for removal the shadow or other noises, first-layer is the object removal, and second-layer is the noises removal.
This two-layer is shown in Fig. 6.

And the removal results are shown in Fig. 7.
From the results, we can see that a high precision salient object detection results can reduce excessive expansion area and save erosion operations’ time in removal algorithm.
\subsection{Scene Reconstruction}
\label{sec:4.6}
After getting the segmented objects and background sceneries we are interested in, then we can reconstruct the scenes as we like.
We reconstruct four scenes which are shown in Fig. 8. From which we can see that the good performance of the proposed algorithm can lead to an exquisite image montage.
\section{Small Target Detection Application}
\label{sec:5}
In this section, we proposed a novel approach to detect small targets by combing the proposed algorithm.
Our small target detection application is divided into two stages, including dark channel prior location and small target accurate detection.
\begin{figure}[!htb]
\label{fig:9}
\begin{center}
\includegraphics[width=4.7 in,height =2.815in]{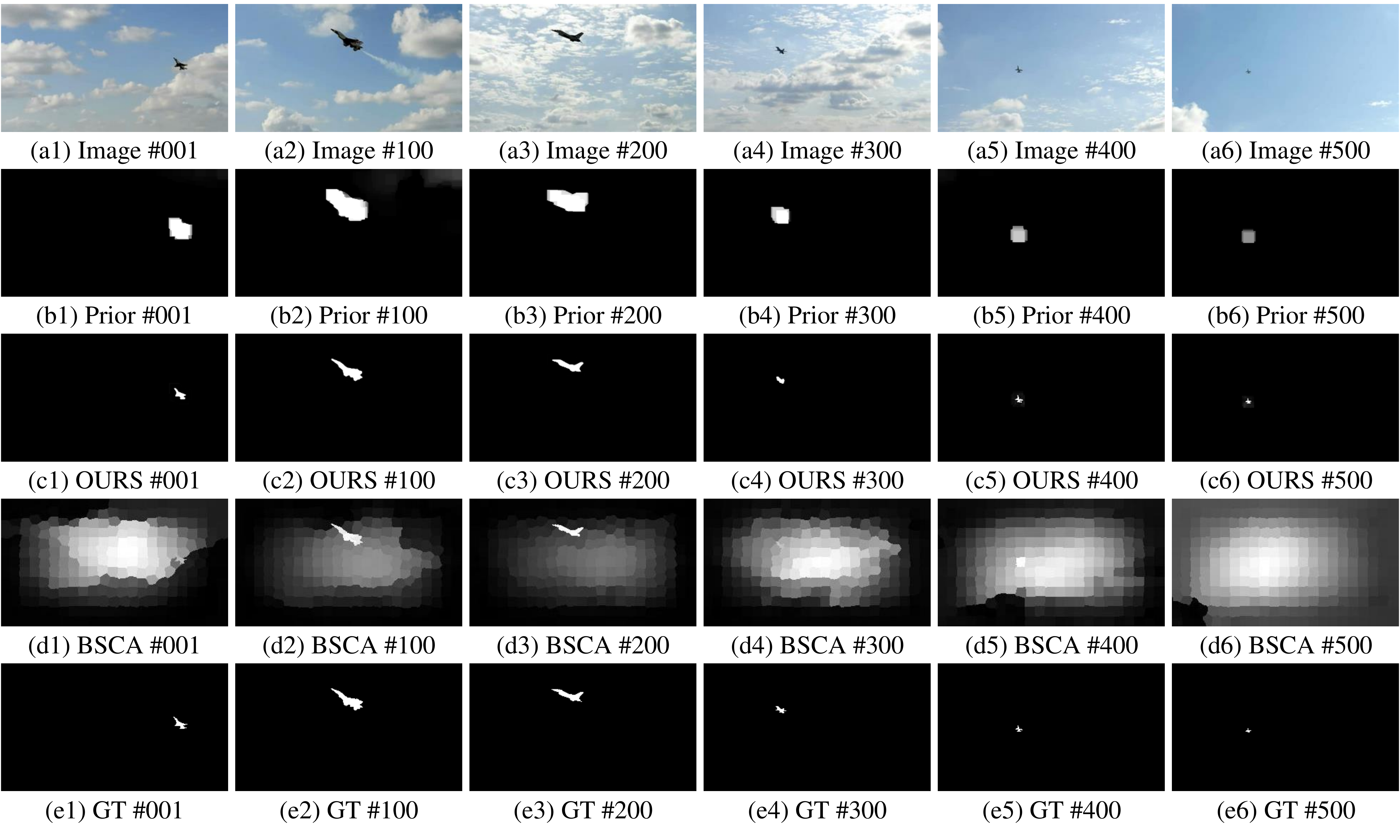}
\end{center}
\caption{The small target detection results. (a1)-(a6) represent different frames of original video.(b1)-(b6) represent different frames of the dark channel prior location results.(c1)-(c6) represent different frames of the proposed algorithm combined with dark channel prior detection results. (d1)-(d6) represent different frames of the BSCA method~\cite{Qin2015Saliency}. (e1)-(e6) represent different frames of the ground truth. }
\label{fig:short}
\end{figure}
\subsection{Dark Channel Prior Location}
\label{sec:5.1}
The dark channel prior is a popular prior which is widely used
in image haze removal field. It is based on the statistics of outdoor haze-free images.
The dark channel can detect the most haze-opaque region and improve the atmospheric light
estimation. Inspired by dark channel prior~\cite{He2009Single}, we find that the foreground and background have
the different transmissivity, so, we can distinguish the foreground objects from the backgrounds. We combine this theory and the proposed saliency detection algorithm to small target detection fields. And we denote the results map of dark channel prior as $D_{cp}$.
\subsection{Small Target Accurate Detection}
\label{sec:5.2}
In this stage, we combine the proposed algorithm and dark channel prior to detect the small target. First, we use the dark channel prior $D_{cp}$ to replace the depth map $I_d$.
Then, we use the proposed algorithm to detect the small target. The experimental results on the small target detection datasets~\cite{Lou2016Small} by applying the proposed algorithm to small target detections， which are shown in Fig. 9. From the comparison, we can get that our detection results are better than the other methods, such as BSCA~\cite{Qin2015Saliency}.
\section{Conclusion}
\label{sec:7}
In this paper, we proposed a multilayer backpropagation saliency detection algorithm based on depth mining. First, we get the additional cues from the preprocessing layer. Then, the proposed algorithm exploits depth cue information of three layers: in the first layer, we mix depth cue to prominent salient object; in the second layer, we extend depth map to prominent salient object’ edges; in the third layer, we reprocess depth cue to eliminate background noises.  And the experiments’ results show that the proposed method outperforms the existing algorithms in both accuracy and robustness in different scenarios. Besides, we demonstrate an innovative application of our algorithm to image montage and the experimental results show that a precision salient object detection can lead to a fine image montage. At last, we give a novel approach to the small target detection application. To encourage future work, we make the source codes and other related materials public. All these can be found on our project website.

\end{document}